%% file: paper.tex
\documentclass[10pt,twocolumn,letterpaper]{article}
\usepackage{cvpr}
\usepackage{times}
\usepackage{epsfig}
\usepackage{graphicx}
\usepackage{amsmath}
\usepackage{amssymb}
\input{math_commands.tex}

\usepackage{balance}
\usepackage{multirow}
\usepackage{booktabs}
\usepackage{subcaption}
\usepackage{enumitem}
\usepackage{pifont}
\usepackage{makecell}
\usepackage{nopageno}
\cvprfinalcopy
\ifcvprfinal\fi
\graphicspath{img}
\newcommand{\cmark}{\ding{51}}%
\newcommand{\xmark}{\ding{55}}%
\title{Conditional Channel Gated Networks for Task-Aware Continual Learning\vspace{-0.2cm}}
\author{
Davide Abati$^1$\thanks{Research conducted during an internship at Qualcomm Technologies Netherlands B.V.} \qquad Jakub Tomczak$^2$ \qquad Tijmen Blankevoort$^2$ \qquad Simone Calderara$^1$ \\ Rita Cucchiara$^1$ \qquad Babak Ehteshami Bejnordi$^2$\vspace{-0.3cm}\\\\
\vspace{-0.3cm}
\begin{tabular}{cc}
$^1$University of Modena and Reggio Emilia 
& \makecell{$^2$Qualcomm AI Research\thanks{Qualcomm AI Research is an initiative of Qualcomm Technologies, Inc.}\\
Qualcomm Technologies Netherlands B.V.}\\
\small{\texttt{\{name.surname\}@unimore.it}} & \small{\texttt{\{jtomczak,tijmen,behtesha\}@qti.qualcomm.com}}
\end{tabular}
}
\newcommand{\res}[2]{$#1$ \footnotesize{$\pm#2$}}
\begin{document}
\maketitle
\begin{abstract}
\input{sections/abstract.tex}
\end{abstract}
\setlength\parindent{0pt}
\input{sections/introduction.tex}
\input{sections/related.tex}
\input{sections/model.tex}
\input{sections/experiments.tex}
\input{sections/conclusion.tex}
\clearpage
{\small
\balance
\bibliographystyle{ieee_fullname}
\bibliography{refs}
}
\clearpage
\Large{\textbf{Supplementary Material}}
\normalsize
\setcounter{section}{0}
\input{supplementary/sections/implementation_details.tex}
\input{supplementary/sections/wgan_details.tex}
\input{supplementary/sections/numbers_for_figures.tex}
\input{supplementary/sections/conditional_generators.tex}
\input{supplementary/sections/reproducibility.tex}
\input{supplementary/floats/wgan_table.tex}
\end{document}

%% file: math_commands.tex
\usepackage{amsmath,amsfonts,bm}

\def\eqref#1{equation~\ref{#1}}
\def\1{\bm{1}}

\DeclareMathAlphabet{\mathsfit}{\encodingdefault}{\sfdefault}{m}{sl}
\SetMathAlphabet{\mathsfit}{bold}{\encodingdefault}{\sfdefault}{bx}{n}

\newcommand{\E}{\mathbb{E}}

\newcommand{\R}{\mathbb{R}}

%% file: sections/abstract.tex
\vspace{-0.2cm}
Convolutional Neural Networks experience catastrophic forgetting when optimized on a sequence of learning problems: as they meet the objective of the current training examples, 
their performance on previous tasks drops drastically.
In this work, we introduce a novel framework to tackle this problem with conditional computation.
We equip each convolutional layer with task-specific gating modules, selecting which filters to apply on the given input. This way, we achieve two appealing properties. 
Firstly, the execution patterns of the gates allow to identify and protect important filters, ensuring no loss in the performance of the model for previously learned tasks. Secondly, by using a sparsity objective, we can promote the selection of a limited set of kernels, allowing to retain sufficient model capacity to digest new tasks.
Existing solutions require, at test time, awareness of the task to which each example belongs to. This knowledge, however, may not be available in many practical scenarios.
Therefore, we additionally introduce a task classifier that predicts the task label of each example, to deal with settings in which a task oracle is not available.
We validate our proposal on four continual learning datasets. Results show that our model consistently outperforms existing methods both in the presence and the absence of a task oracle. Notably, on Split SVHN and Imagenet-50 datasets, our model yields up to 23.98\% and 17.42\% improvement in accuracy w.r.t. competing methods.

%% file: sections/introduction.tex
\vspace{-0.2cm}
\section{Introduction}
Machine learning and deep learning models are typically trained offline, by sampling examples independently from the distribution they are expected to deal with at test time.
However, when trained online in real-world settings, models may encounter multiple tasks as a sequential stream of activities, without having any knowledge about their relationship or duration in time. Such challenges typically arise in robotics~\cite{aljundi2019task}, reinforcement learning~\cite{ring1997child}, vision systems~\cite{ostapenko2019learning} and many more (cf.~Chapter~4 in~\cite{chen2018lifelong}). In such scenarios, deep learning models suffer from \textit{catastrophic forgetting}~\cite{mccloskey1989catastrophic,french1999catastrophic}, meaning they discard previously acquired knowledge to fit the current observations. The underlying reason is that, while learning the new task, models overwrite the parameters that were critical for previous tasks.

Continual learning research (also called \textit{lifelong} or \textit{incremental} learning) tackles the above mentioned issues~\cite{chen2018lifelong}.
The typical setting considered in the literature is that of a model learning disjoint classification problems one-by-one.
Depending on the application requirements, the task for which the current input should be analyzed may or may not be known. The majority of the methods in the literature assume that the label of the task is provided during inference. Such a continual learning setting is generally referred to as task-incremental. 
In many real-world applications, such as classification and anomaly detection systems, 
a model can seamlessly instantiate a new task whenever novel classes emerge from the training stream.
However, once deployed in the wild, it has to process inputs without knowing in which training task similar observations were encountered.
Such a setting, in which task labels are available only during training, 
is known as class-incremental~\cite{threescenarios}. 
Existing methods employ different strategies to mitigate catastrophic forgetting, such as memory buffers~\cite{icarl,gem}, knowledge distillation~\cite{lwf}, synaptic consolidation~\cite{ewc} and parameters masking~\cite{packnet,hat}. 
However, recent evidence has shown that existing solutions fail, even for simple datasets, whenever task labels are not available at test time~\cite{threescenarios}.

This paper introduces a solution based on conditional-computing to tackle both task-incremental and class-incremental learning problems. 
Specifically, our framework relies on separate task-specific classification heads (\textit{multi-head} architecture), and it employs channel-gating~\cite{gaternet,babak} in every layer of the (shared) feature extractor.
To this aim, we introduce task-dedicated gating modules that dynamically select which filters to apply conditioned on the input feature map. 
Along with a sparsity objective encouraging the use of fewer units, this strategy enables per-sample model selection and can be easily queried for information about which weights are essential for the current task.
Those weights are frozen when learning new tasks, but gating modules can dynamically select to either use or discard them. 
Contrarily, units that are never used by previous tasks are reinitialized and made available for acquiring novel concepts.
This procedure prevents any forgetting of past tasks and allows considerable computational savings in the forward propagation.

Moreover, we obviate the need for a task label during inference by introducing a task classifier selecting which classification head should be queried for the class prediction.
We train the task classifier alongside the classification heads under the same incremental learning constraints. 
To mitigate forgetting on the task classification side, we rely on example replay from either episodic or generative memories. 
In both cases, we show the benefits of performing rehearsal at a task-level, as opposed to previous replay methods that operate at a class-level~\cite{icarl,agem}.
To the best of our knowledge, this is the first work that carries out supervised task prediction in a class-incremental learning setting.

We perform extensive experiments on four datasets of increasing difficulty, both in the presence and absence of a task oracle at test time. Our results show that, whenever task labels are available, our model effectively prevents the forgetting problem, and performs similarly to or better than state-of-the-art solutions. In the task agnostic setting, we consistently outperform competing methods.

%% file: sections/related.tex
\section{Related work}
\textbf{Continual learning.}
Catastrophic forgetting has been a well-known problem of neural networks ~\cite{mccloskey1989catastrophic}. Early approaches to alleviate the issue involved orthogonal representation learning and replay of prior samples~\cite{french1999catastrophic}. The recent advent in deep learning has led to the widespread use of deep neural networks in the continual learning field. First attempts, such as Progressive Neural Networks~\cite{progressivenns} tackle the forgetting problem by introducing a new set of parameters for each new task at the expense of limited scalability. Another popular solution is to apply knowledge distillation by using the past parametrizations of the model as a reference when learning new tasks~\cite{lwf}.

\textit{Consolidation} approaches emerged recently with the focus of identifying the weights that are critically important for prior tasks and preventing significant updates to them during the learning of new tasks. 
The relevance/importance estimation for each parameter can be carried out through the Fisher Information Matrix~\cite{ewc}, the path integral of loss gradients~\cite{si}, gradient magnitude~\cite{mas} and a posteriori uncertainty estimation in a Bayesian Neural Network~\cite{vcl}.

Other popular consolidation strategies rely on the estimation of binary masks that directly map each task to the set of parameters responsible for it. Such masks can be estimated either by random assignment~\cite{xdg}, pruning~\cite{packnet} or gradient descent~\cite{piggyback,hat}. However, existing mask-based approaches can only operate in the presence of an oracle providing the task label. 
Our work is akin to the above-mentioned models, with two fundamental differences: i) our binary masks (gates) are dynamically generated and depend on the network input, and ii) we promote mask-based approaches to class-incremental learning settings, by relying on a novel architecture comprising a task classifier.

Several models allow access to a finite-capacity memory buffer (\textit{episodic} memory), holding examples from prior tasks. A popular approach is iCaRL~\cite{icarl}, which computes class prototypes as the mean feature representation of stored memories, and classifies test examples in a nearest-neighbor fashion. Alternatively, other approaches intervene in the training algorithm, proposing to adjust the gradient computed on the current batch towards an update direction that guarantees non-destructive effects on the stored examples~\cite{gem,agem,mer}. Such an objective can imply the formalization of constrained optimization problems~\cite{gem,agem} or the employment of meta-learning algorithms~\cite{mer}. Differently, \textit{generative} memories do not rely on the replay of any real example whatsoever, in favor of generative models from which fake examples of past tasks can be efficiently sampled~\cite{shin2017continual,wu2018memory,ostapenko2019learning}. 
In this work, we also rely on either episodic or generative memories to deal with the class-incremental learning setting. However, we carry out replay only to prevent forgetting of the task predictor, thus avoiding to update task-specific classification heads.
\\\\
\textbf{Conditional computation.}
Conditional computation research focuses on deep neural networks that adapt their architecture to the given input. Although the first work has been applied to language modeling~\cite{moe}, several works applied such concept to computer vision problems. In this respect, prior works employ binary gates deciding whether a computational block has to be executed or skipped. 
Such gates may either drop entire residual blocks~\cite{aig,skipnet} or specific units within a layer~\cite{gaternet,babak}.
In our work, we rely on the latter strategy, learning a set of task-specific gating modules selecting which kernels to apply on the given input. To our knowledge, this is the first application of data-dependent channel-gating in continual learning.

%% file: sections/model.tex
\section{Model}
\def\x{{\bf x}}
\def\y{{\bf y}}
\def\t{{\bf t}}
\def\p{p_{\theta}}
\subsection{Problem setting and objective}
We are given a parametric model, i.e., a neural network, called a \textit{backbone} or \textit{learner} network, which is exposed to a sequence of $N$ tasks to be learned, $\mathcal{T} = \{T_1, \dots, T_N\}$. Each task $T_i$ takes the form of a classification problem, $T_i = \{\x_j, y_j\}_{j=1}^{n_i}$, where $\x_j \in \mathbb{R}^m$ and $y_j \in \{1,\dots,C_i\}$.

A task-incremental setting requires to optimize:
\begin{equation}
\label{eq:cl_multihead}
    \max_{\theta} \quad \E_{\t\sim \mathcal{T}}\left[\E_{(\x, \y)\sim T_\t}\left[\log \p(\y|\x,\t)\right]\right],
\end{equation}
where $\theta$ identifies the parametrization of the learner network, and $\x$, $\y$ and $\t$ are random variables associated with the observation, the label and the task of each example, respectively.
Such a maximization problem is subject to the continual learning constraints: as the model observes tasks sequentially, the outer expectation in Eq.~\ref{eq:cl_multihead} is troublesome to compute or approximate. Notably, this setting requires the assumption that the identity of the task each example belongs to is known at both training and test stages. Such information can be exploited in practice to isolate relevant output units of the classifier, preventing the competition between classes belonging to different tasks through the same softmax layer (\textit{multi-head}).

Class-incremental models solve the following optimization:
\begin{equation}
\label{eq:cl_singlehead}
    \max_{\theta} \quad \E_{\t\sim \mathcal{T}}\left[\E_{(\x, \y)\sim T_\t}\left[\log \p(\y|\x)\right]\right].
\end{equation}
Here, the absence of task conditioning prevents any form of task-aware reasoning in the model. This setting requires to merge the output units into a single classifier (\textit{single-head}) in which classes from different tasks compete with each other, often resulting in more severe forgetting~\cite{threescenarios}. Although the model could learn based on task information, this information is not available during inference.\\

To deal with observations from unknown tasks, while retaining advantages of multi-head settings, we will jointly optimize for class as well as task prediction, as follows: 
\begin{align}
\label{eq:our_objective}
    \nonumber
    \max_{\theta} \quad&\E_{\t\sim \mathcal{T}}\left[\E_{(\x, \y)\sim T_\t}\left[\log \p(\y, \t|\x)\right]\right] = \\
    &\E_{\t\sim \mathcal{T}}\left[\E_{(\x, \y)\sim T_\t}\left[\log \p(\y|\x,\t) + \log \p(\t|\x)\right]\right].
\end{align}
Eq.~\ref{eq:our_objective} describes a twofold objective. 
On the one hand, the term $\log p(\y|\x,\t)$ is responsible for the \textit{class classification given the task}, and resembles the multi-head objective in Eq.~\ref{eq:cl_multihead}. 
On the other hand, the term $\log p(\t|\x)$ aims at \textit{predicting the task} from the observation. This prediction relies on a task classifier, which is trained incrementally in a single-head fashion. 
Notably, the objective in Eq.~\ref{eq:our_objective} shifts the single-head complexities from a class prediction to a task prediction level, with the following benefits:
\begin{itemize}[noitemsep]
    \item given the task label, there is no drop in class prediction accuracy;
    \item classes from different tasks never compete with each other, neither during training nor during test;
    \item the challenging single-head prediction step is shifted from class to task level; as tasks and classes form a two-level hierarchy, the prediction of the former is arguably easier (as it acts at a coarser semantic level).
\end{itemize}
\subsection{Multi-head learning of class labels}
\label{sec:class_classification}
\def\hi{{\bf h}^l}
\def\ho{{\bf h}^{l+1}}
\def\hgo{\hat{{\bf h}}^{l+1}}
\newcommand{\G}[2]{G^{#1}_{#2}}
\def\Gtl{\G{l}{t}}
\input{floats/gates_figure.tex}
\input{floats/model_figure.tex}
In this section, we introduce the conditional computation model we used in our work. Fig.~\ref{fig:gating} illustrates the gating mechanism used in our framework. We limit the discussion of the gating mechanism to the case of convolutional layers, as it also applies to other parametrized mappings such as fully connected layers or residual blocks. Consider $\hi \in \R^{c_{in}^l,h,w}$ and $\ho\in \R^{c_{out}^l,h',w'}$ to be the input and output feature maps of the $l$-th convolutional layer respectively.
Instead of $\ho$, we will forward to the following layer a sparse feature map $\hgo$, obtained by pruning uninformative channels. During the training of task $t$, the decision regarding which channels have to be activated is delegated to a gating module $\Gtl$, that is conditioned on the input feature map $\hi$:
\begin{equation}
    \hgo = \Gtl(\hi) \odot \ho,
\end{equation}
where $\Gtl(\hi) = [g_1^l,\dots,g_{c_{out}^{l}}^l]$, $g_i^l\in\{0,1\}$, and $\odot$ refers to channel-wise multiplication.
To be compliant with the incremental setting, we instantiate a new gating module each time the model observes examples from a new task. However, each module is designed as a light-weight network with negligible computation costs and number of parameters. Specifically, each gating module comprises a Multi-Layer Perceptron (MLP) with a single hidden layer featuring 16 units, followed by a batch normalization layer~\cite{batchnorm} and a ReLU activation. A final linear map provides log-probabilities for each output channel of the convolution.

Back-propagating gradients through the gates is challenging, as non-differentiable thresholds are employed to take binary on/off decisions. Therefore, we rely on the Gumbel-Softmax sampling~\cite{jang2016categorical,maddison2016concrete}, and get a biased estimate of the gradient utilizing the straight-through estimator~\cite{bengio2013estimating}. Specifically, we employ the hard threshold in the forward pass (zero-centered) and the sigmoid function in the backward pass (with temperature $\tau=2/3$).

Moreover, we penalize the number of active convolutional kernels with the sparsity objective:
\begin{equation}
\label{eq:sparsity}
\mathcal{L}_{sparse}=\E_{(\x, \y)\sim T_\t}\left[\frac{\lambda_s}{L}\sum_{l=1}^L\frac{\|\Gtl(\hi)\|_1}{c_{out}^l}\right],
\end{equation}
where $L$ is the total number of gated layers, and $\lambda_s$ is a coefficient controlling the level of sparsity. The sparsity objective instructs each gating module to select a minimal set of kernels, allowing us to conserve filters for the optimization of future tasks. Moreover, it allows us to effectively adapt the capacity of the allocated network depending on the difficulty of the task and the observation at hand. Such a data-driven model selection contrasts with other continual learning strategies that employ fixed ratios for model growing~\cite{progressivenns} or weight pruning~\cite{packnet}.

At the end of the optimization for task $t$, we compute a relevance score $r_k^l$ for each unit in the $l$-th layer by estimating the firing probability of their gates on a validation set $T_\t^{val}$:
\begin{equation}
r_k^{l,t} = \E_{(\x, \y)\sim T_\t^{val}}[p(\mathbb{I}[g_k^l = 1])],
\end{equation}
where $\mathbb{I}[\cdot]$ is an indicator function, and $p(\cdot)$ denotes a probability distribution.
By thresholding such scores, we obtain two sets of kernels. On the one hand, we \textit{freeze relevant kernels} for the task $t$, so that they will be available but not updatable during future tasks. On the other hand, we \textit{re-initialize non-relevant kernels}, and leave them learnable by subsequent tasks. In all our experiments, we use a threshold equal to 0, which prevents any forgetting at the expense of a reduced model capacity left for future tasks.

Note that within this framework, it is trivial to monitor the number of learnable units left in each layer. As such, if the capacity of the backbone model saturates, we can quickly grow the network to digest new tasks. However, because the gating modules of new tasks can dynamically choose to use previously learned filters (if relevant for their input), learning of new tasks generally requires less learnable units. In practice, we never experienced the saturation of the backbone model for learning new tasks. Apart from that, because of our conditional channel-gated network design, increasing the model capacity for future tasks will have minimal effects on the computation cost at inference, as reported by the analysis in Sec.~\ref{sec:model_analysis}.
\subsection{Single-head learning of task labels}
\label{sec:task_classification}
\newcommand{\hhat}[2]{\hat{{\bf h}}^{#1}_{#2}}
\def\hcat{\\h}
The gating scheme presented in Sec.~\ref{sec:class_classification} allows the immediate identification of important kernels for each past task. However, it cannot be applied in the task-agnostic setting as is, since it requires the knowledge about which gating module $\G{l}{x}$ has to be applied for layer $l$, where $x\in \{1,\dots,t\}$ represents the unknown task.
Our solution is to employ all gating modules $[\G{l}{1},\dots,\G{l}{t}]$, and to propagate all gated layer outputs $[\hhat{l+1}{1},\dots,\hhat{l+1}{t}]$ forward.
In turn, the following layer $l+1$ receives the list of gated outputs from layer $l$, applies its gating modules $[\G{l+1}{1},\dots,\G{l+1}{t}]$ and yields the list of outputs $[\hhat{l+2}{1},\dots,\hhat{l+2}{t}]$.
This mechanism generates parallel streams of computation in the network, sharing the same layers but selecting different sets of units to activate for each of them (Fig.~\ref{fig:model}).
Despite the fact that the number of parallel streams grows with the number of tasks, we found our solution to be computationally cheaper than the backbone network (see Sec.~\ref{sec:model_analysis}). This is because of the gating modules which select a limited number of convolutional filters in each stream.

After the last convolutional layer, indexed by $L$, we are given a list of $t$ candidate feature maps $[\hhat{L+1}{1},\dots,\hhat{L+1}{t}]$ and as many classification heads. The task classifier is fed with a concatenation of all feature maps:
\begin{equation}
\hcat=\bigoplus\limits_{i=1}^{t}[\mu(\hhat{L+1}{i})],
\end{equation}
where $\mu$ denotes the global average pooling operator over the spatial dimensions and $\bigoplus$ describes the concatenation along the feature axis. 
The architecture of the task classifier is based on a shallow MLP with one hidden layer featuring 64 ReLU units, followed by a softmax layer predicting the task label. We use the standard cross-entropy objective to train the task classifier. Optimization is carried out jointly with the learning of class labels at task $t$. Thus, the network not only learns features to discriminate the classes inside task $t$, but also to allow easier discrimination of input data from task $t$ against all prior tasks. 

The single-head task classifier is exposed to catastrophic forgetting. Recent papers have shown that replay-based strategies represent the most effective continual learning strategy in single-head settings~\cite{threescenarios}. Therefore, we choose to ameliorate the problem by rehearsal. In particular, we consider the following approaches.
\\
\\
\textbf{Episodic memory.} A small subset of examples from prior tasks is used to rehearse the task classifier. During the training of task $t$, the buffer holds $C$ random examples from past tasks $1, \dots, t-1$ (where $C$ denotes a fixed capacity). Examples from the buffer and the current batch (from task $t$) are re-sampled so that the distribution of task labels in the rehearsal batch is uniform. At the end of task $t$, the data in the buffer is subsampled so that each past task holds $m=C/t$ examples. Finally, $m$ random examples from task $t$ are selected for storage.
\\
\\
\textbf{Generative memory.} A generative model is employed for sampling fake data from prior tasks. Specifically, we utilize Wasserstein GANs with Gradient Penalty (WGAN-GP~\cite{wgangp}). To overcome forgetting in the sampling procedure, we use multiple generators, each of which models the distribution of examples of a specific task.
\\
\\
In both cases, replay is only employed for rehearsing the task classifier and not the classification heads. To summarize, the complete objective of our model includes: the cross-entropy at a class level ($\p(\y|\x,\t)$ in Eq.~\ref{eq:our_objective}), the cross-entropy at a task level ($\p(\t|\x)$ in Eq.~\ref{eq:our_objective}) and the sparsity term ($\mathcal{L}_{sparse}$ in Eq.~\ref{eq:sparsity}).

%% file: floats/gates_figure.tex
\begin{figure}
\centering
\includegraphics[width=0.9\columnwidth]{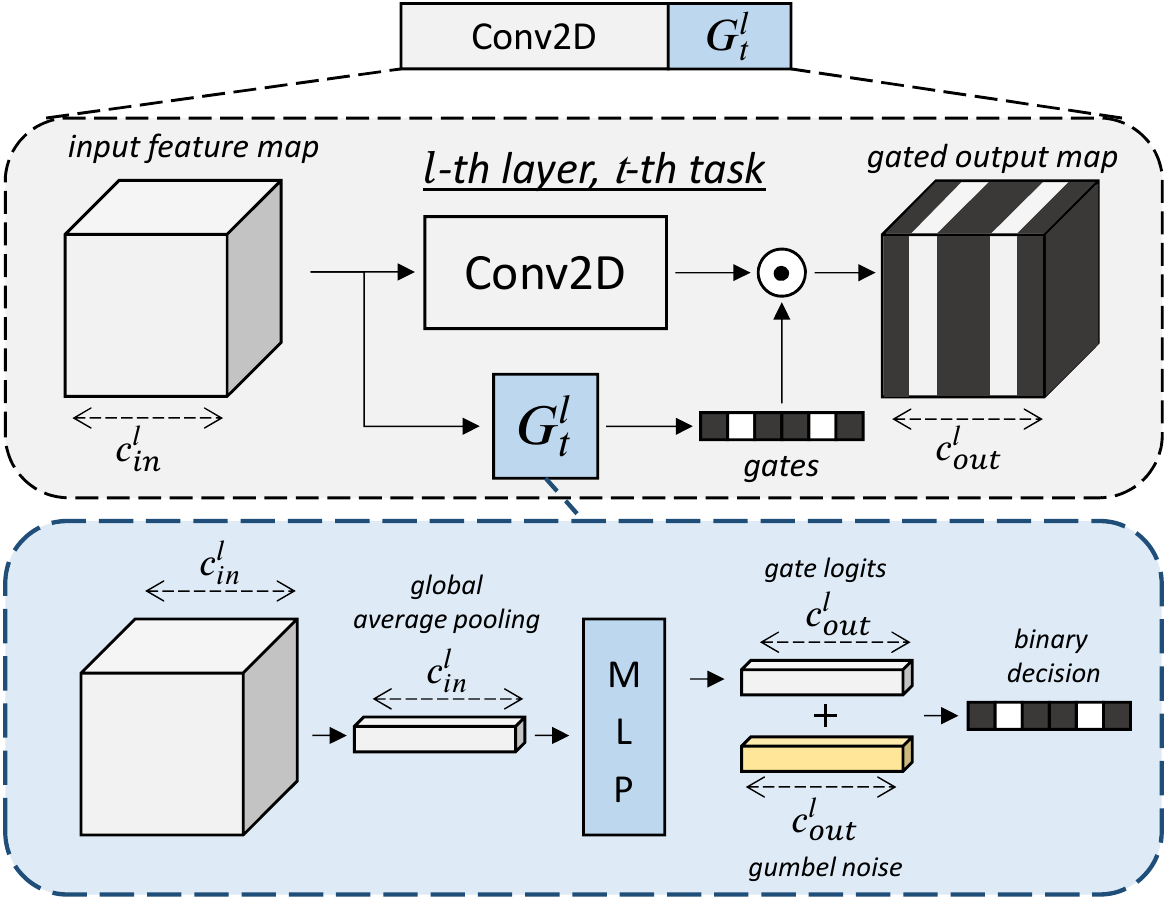}
\caption{The proposed gating scheme for a convolution layer. Depending on the input feature map, the gating module $\G{l}{t}$ decides which kernels should be used.}
\label{fig:gating}
\end{figure}

%% file: floats/model_figure.tex
\begin{figure*}[tb]
    \centering
    \includegraphics[width=0.9\textwidth]{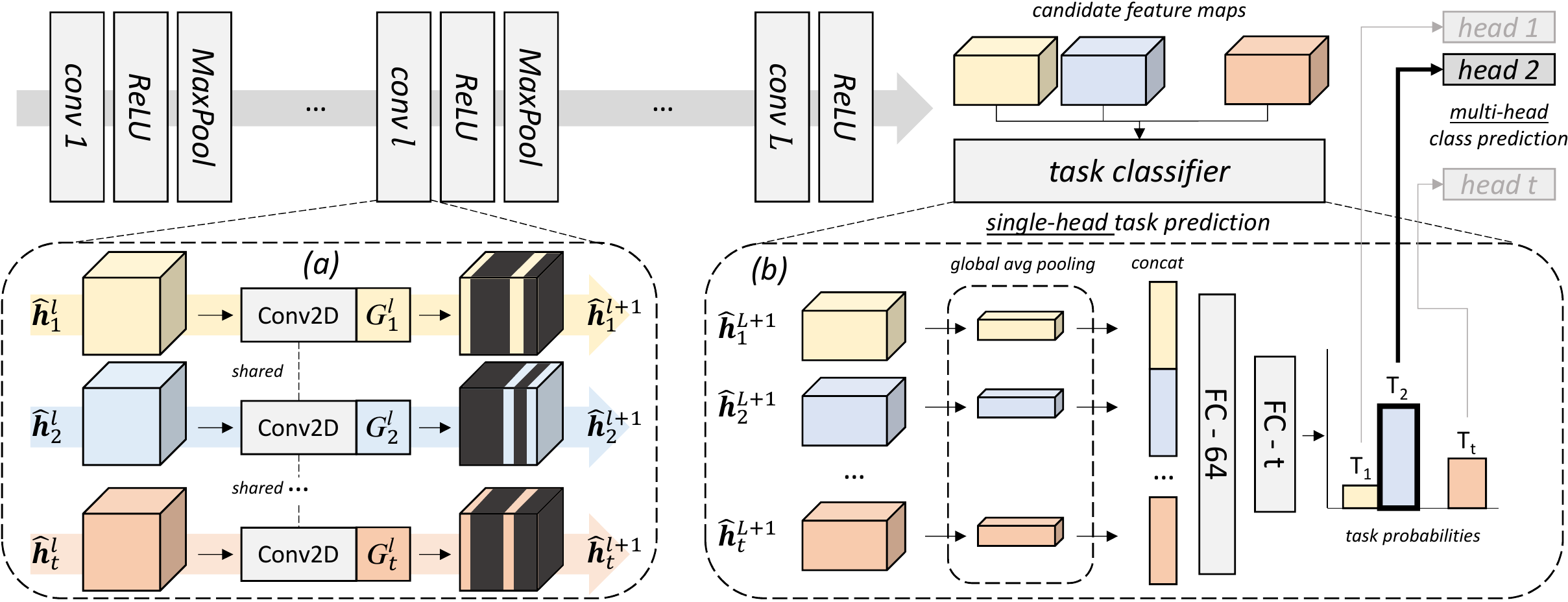}
    \caption{Illustration of the task prediction mechanism for a generic backbone architecture. First (block `a'), the $l$-th convolutional layer is fed with multiple gated feature maps, each of which is relevant for a specific task. Every feature map is then convolved with kernels selected by the corresponding gating module $\G{l}{x}$, and forwarded to the next module. At the end of the network the task classifier (block `b') takes as input candidate feature maps and decides which task to solve.}
    \label{fig:model}
\end{figure*}

%% file: sections/experiments.tex
\input{floats/resnet_gates_figure.tex}
\input{floats/task_oracle_table.tex}
\section{Experiments}
\label{sec:experiments}
\subsection{Datasets and backbone architectures}
\label{sec:datasets}
We experiment with the following datasets:
\begin{itemize}[noitemsep]
    \item Split MNIST: the MNIST handwritten classification benchmark~\cite{mnist} is split into 5 subsets of consecutive classes. This results into 5 binary classification tasks that are observed sequentially.
    \item Split SVHN: the same protocol applied as in Split MNIST, but employing the SVHN dataset~\cite{svhn}.
    \item Split CIFAR-10: the same protocol applied as in Split MNIST, but employing the CIFAR-10 dataset~\cite{cifar}.
    \item Imagenet-50~\cite{ostapenko2019learning}: a subset of the iILSVRC-2012 dataset~\cite{imagenet} containing 50 randomly sampled classes and 1300 images per category, split into 5 consecutive 10-way classification problems. Images are resized to a resolution of 32x32 pixels.
\end{itemize}
As for the backbone models, for the MNIST and SVHN benchmarks, we employ a three-layer CNN with 100 filters per layer and ReLU activations (\textit{SimpleCNN} in what follows). All convolutions except for the last one are followed by a 2x2 max-pooling layer. Gating is applied after the pooling layer. A final global average pooling followed by a linear classifier yields class predictions. For the \mbox{CIFAR-10} and Imagenet-50 benchmarks we employed a ResNet-18~\cite{resnet} model as backbone. The gated version of a ResNet basic block is represented in Fig.~\ref{fig:gated_resnet}. As illustrated, two independent sets of gates are applied after the first convolution and after the residual connection, respectively.

All models were trained with SGD with momentum until convergence. After each task, model selection is performed for all models by monitoring the corresponding objective on a held-out set of examples from the current task (i.e., we don't rely on examples of past tasks for validation purposes). 
We apply the sparsity objective introduced in Sec.~\ref{sec:class_classification} only after a predetermined number of epochs, to provide the model the possibility to learn meaningful kernels before starting pruning the uninformative ones.
We refer to the supplementary material for further implementation details.
\subsection{Task-incremental setting}
In the task-incremental setting, an oracle can be queried for task labels during test time. Therefore, we don't rely on the task classifier, exploiting ground-truth task labels to select which gating modules and classification head should be active. This section validates the suitability of the proposed data-dependent gating scheme for continual learning.
We compare our model against several competing methods:
\begin{itemize}[noitemsep]
    \item[--] \textit{Joint}: the backbone model trained jointly on all tasks while having access to the entire dataset. We considered its performance as the upper bound.
    \item[--] \textit{Ewc-On}~\cite{progressandcompress}: the online version of Elastic Weight Consolidation, relying on the latest MAP estimate of the parameters and a running sum of Fisher matrices.
    \item[--] \textit{LwF}~\cite{lwf}: an approach in which the task loss is regularized by a distillation objective, employing the initial state of the model on the current task as a teacher.
    \item[--] \textit{HAT}~\cite{hat}: a mask-based model conditioning the active units in the network on the task label. Despite being the most similar approach to our method, it can only be applied in task-incremental settings.
\end{itemize}
Tab.~\ref{tab:task_oracle} reports the comparison between methods, in terms of accuracy on all tasks after the whole training procedure.

Despite performing very similarily for MNIST, the gap in the consolidation capability of different models emerges as the dataset grows more and more challenging. It is worth mentioning several recurring patterns. 
First, LwF struggles when the number of tasks grows larger than two. Although its distillation objective is an excellent regularizer against forgetting, it does not allow enough flexibility to the model to acquire new knowledge. Consequently, its accuracy on the most recent task gradually decreases during sequential learning, whereas the performance on the first task is kept very high.
Moreover, results highlight the suitability of gating-based schemes (HAT and ours) with respect to other consolidation strategies such as EWC Online. Whereas the former ones prevent any update of relevant parameters, the latter approach only penalizes updating them, eventually incurring a significant degree of forgetting. Finally, the table shows that our model either performs on-par or outperforms HAT on all datasets, suggesting the beneficial effect of our data-dependent gating scheme and sparsity objective.
\subsection{Class-incremental with episodic memory}
Next, we move to a class-incremental setting in which no awareness of task labels is available at test time, significantly increasing the difficulty of the continual learning problem. In this section, we set up an experiment for which the storage of a limited amount of examples (buffer) is allowed.
We compare against:
\begin{itemize}[noitemsep]
    \item[--] Full replay: upper bound performance given by replay to the network of an unlimited number of examples.
    \item[--] iCaRL~\cite{icarl} an approach based on a nearest-neighbor classifier exploiting examples in the buffer. We report the performances both with the original buffer-filling strategy (\textit{iCaRL-mean}) and with the randomized algorithm used for our model (\textit{iCaRL-rand});
    \item[--] A-GEM~\cite{agem}: a buffer-based method correcting parameter updates on the current task so that they don't contradict the gradient computed on the stored examples.
\end{itemize}
Results are summarized in Fig.~\ref{fig:episodic memory_plots}, illustrating the final average accuracy on all tasks at different buffer sizes for the class-incremental Split-MNIST and Split-SVHN benchmarks. The figure highlights several findings. 
Surprisingly, A-GEM yields a very low performance on MNIST, while providing higher results on SVHN. Further examination on the former dataset revealed that it consistently reaches competitive accuracy on the most recent task, while mostly forgetting the prior ones. 
The performance of iCaRL, on the other hand, does not seem to be significantly affected by changing its buffer filling strategy. Moreover, its accuracy seems not to scale with the number of stored examples.
In contrast to these methods, our model primarily utilizes the few stored examples for the rehearsal of coarse-grained task prediction, while retaining the accuracy of fine-grained class prediction. As shown in Fig.~\ref{fig:episodic memory_plots}, our approach consistently outperforms competing approaches in the class-incremental setting with episodic memory.
\input{floats/episodic_memory_plots.tex}
\subsection{Class-incremental with generative memory}
\label{sec:generative_memory}
Next, we experiment with a class-incremental setting in which no examples are allowed to be stored whatsoever. A popular strategy in this framework is to employ generative models to approximate the distribution of prior tasks and rehearse the backbone network by sampling fake observations from them. Among these, DGM~\cite{ostapenko2019learning} is the state-of-the-art approach, which proposes a class-conditional GAN architecture paired with a hard attention mechanism similar to the one of HAT~\cite{hat}. Fake examples from the GAN generator are replayed to the discriminator, which includes an auxiliary classifier providing a class prediction. As for our model, as mentioned in Sec.~\ref{sec:task_classification}, we rely on multiple task-specific generators.
For a detailed discussion of the architecture of the employed WGANs, we refer the reader to the supplementary material.
Tab.~\ref{tab:generative_replay} compares the results of DGM and our model for the class-incremental setting with generative memory. Once again, our method of exploiting rehearsal for only the task classifier proves beneficial. DGM performs particularly well on Split MNIST, where hallucinated examples are almost indistinguishable from real examples. On the contrary, results suggest that class-conditional rehearsal becomes potentially unrewarding as the complexity of the modeled distribution increases, and the visual quality of generated samples degrades.
\subsection{Model analysis}
\label{sec:model_analysis}
\textbf{Episodic vs. generative memory.}
To understand which rehearsal strategy has to be preferred when dealing with class-incremental learning problems, we raise the following question: What is more beneficial between a limited amount of real examples and a (potentially) unlimited amount of generated examples? To shed light on this matter, we report our models' performances on Split SVHN and Split CIFAR-10 as a function of memory budget.
Specifically, we compute the memory consumption of episodic memories as the cumulative size of the stored examples. As for generative memories, we consider the number of bytes needed to store their parameters (in single-precision floating-point format), discarding the corresponding discriminators as well as inner activations generated in the sampling process.
Fig.~\ref{fig:memory_consumption} presents the result of the analysis. 
As can be seen, the variant of our model relying on memory buffers consistently outperforms its counterpart relying on generative modeling.
In the case of CIFAR-10, the generative replay yields an accuracy comparable with an episodic memory of $\approx1.5$~MBs, which is more than 20 times smaller than its generators. 
The gap between the two strategies shrinks on SVHN, due to the simpler image content resulting in better samples from the generators.
Finally, our method, when based on memory buffers, outperforms the DGMw model~\cite{ostapenko2019learning} on Split-SVHN, albeit requiring 3.6 times less memory.
\input{floats/generative_memory_table.tex}
\input{floats/memory_consumption_figure.tex}
\input{floats/gates_visualization_figure.tex}
\\\\
\textbf{Gate analysis.}
We provide a qualitative analysis of the activation of gates across different tasks in Fig.~\ref{fig:gates_visualization}. Specifically, we use the validation sets of Split MNIST and Imagenet-50 to compute the probability of each gate to be triggered by images from different tasks\footnote{we report such probabilities for specific layers: layer 1 for Split MNIST (Simple CNN), block 5 for Imagenet-50 (ResNet-18).}. The analysis of the figure suggests two pieces of evidence: First, as more tasks are observed, previously learned features are re-used. This pattern shows that the model does not fall into degenerate solutions, e.g., by completely isolating tasks into different sub-networks. On the contrary, our model profitably exploits pieces of knowledge acquired from previous tasks for the optimization of the future ones. 
Moreover, a significant number of gates never fire, suggesting that a considerable portion of the backbone capacity is available for learning even more tasks. Additionally, we showcase how images from different tasks activating the same filters show some resemblance in low-level or semantic features (see the caption for details).
\input{floats/mac_count_table.tex}
\\\\
\textbf{On the cost of inference.}
We next measure the inference cost of our model as the number of tasks increases. Tab.~\ref{tab:mac_counts} reports the average number of multiply-add operations (MAC count) of our model on the test set of Split MNIST and Split CIFAR-10 after learning each task. Moreover, we report the MACs of HAT~\cite{hat} as well as the cost of forward propagation in the backbone network (i.e. the cost of any other competing method mentioned it this section). In the task-incremental setting, our model obtains a meaningful saving in the number of operations, thanks to the data-dependent gating modules selecting only a small subset of filters to apply. In contrast, forward propagation in a class-incremental setting requires as many computational streams as the number of tasks observed so far. However, each of them is extremely cheap as few convolutional units are active. As presented in the table, also in the class-incremental setting, the number of operations never exceeds the cost of forward propagation in the backbone model. The reduction in inference cost is particularly significant for Split CIFAR-10, which is based on a ResNet-18 backbone.\\\\
\textbf{Limitations and future works.}
Training our model can require a lot of GPU memory for bigger backbones.
However, by exploiting the inherent sparsity of activation maps, several optimizations are possible.
Secondly, we expect the task classifier to be susceptible to the degree of semantic separation among tasks.
For instance, a setting where tasks are semantically well-defined, like $T_1 = \{\text{cat,dog}\}$, $T_2 = \{\text{car,truck}\}$ (\textit{animals} / \textit{vehicles}), should favor the task classifier with respect to its transpose $T_1 = \{\text{cat,car}\}$, $T_2 = \{\text{dog,truck}\}$. However, we remark that in our experiments the assigment of classes to tasks is always random. Therefore, our model could perform even better in the presence of coherent tasks.

%% file: floats/resnet_gates_figure.tex
\begin{figure}[b]
    \centering
    \includegraphics[width=\columnwidth]{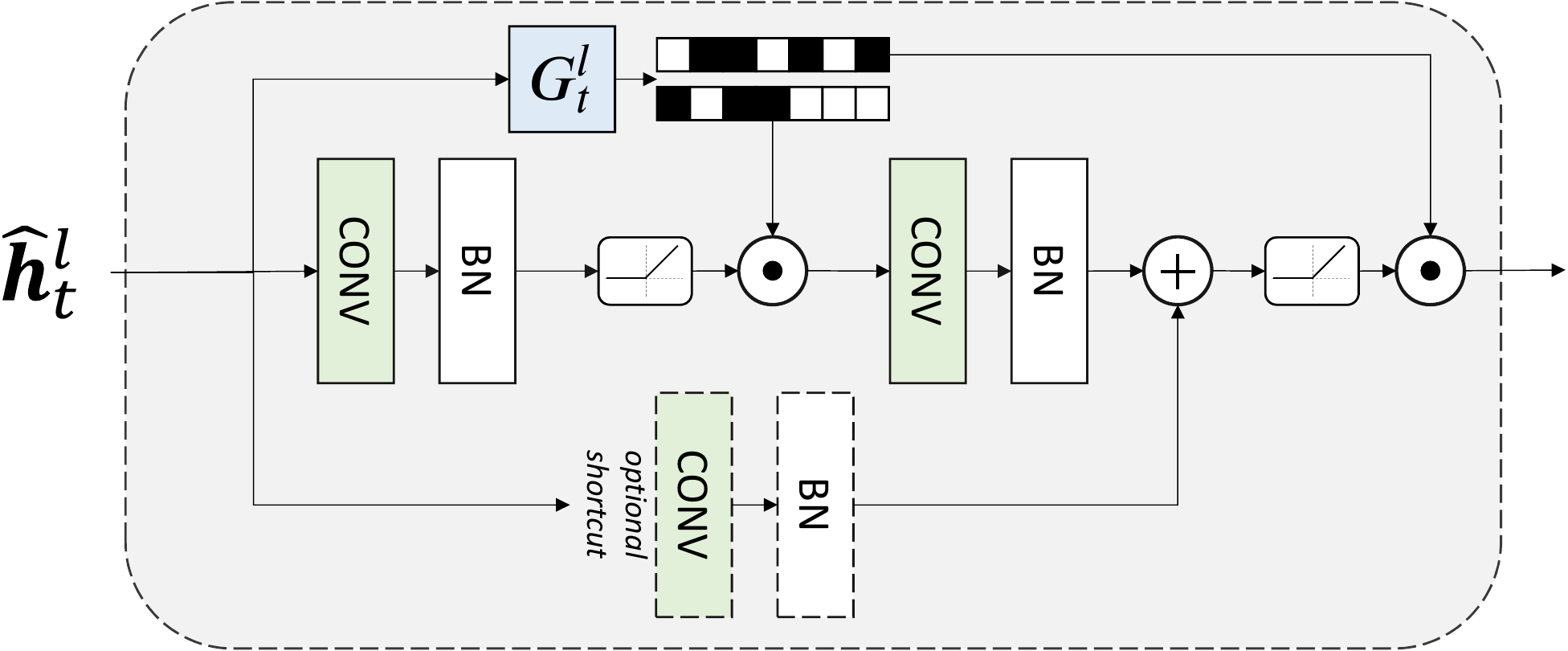}
    \caption{The gating scheme applied to ResNet-18 blocks. Gating on the \textit{shortcut} is only applied when downsampling.}
    \label{fig:gated_resnet}
\end{figure}

%% file: floats/task_oracle_table.tex
\begin{table*}[t]
\centering
\resizebox{\textwidth}{!}{
\begin{tabular}{lcccccccccccccccccccc}
\toprule
&\multicolumn{6}{c}{Split MNIST}
&&\multicolumn{6}{c}{Split SVHN}
&&\multicolumn{6}{c}{Split CIFAR-10}
\\
& $T_1$ & $T_2$ & $T_3$ & $T_4$ & $T_5$ & avg&
& $T_1$ & $T_2$ & $T_3$ & $T_4$ & $T_5$ & avg&
& $T_1$ & $T_2$ & $T_3$ & $T_4$ & $T_5$ & avg
\\\midrule
\textit{Joint (UB)}
&0.999&0.999&0.999&1.000&0.995&0.999&
&0.983&0.972&0.982&0.983&0.941&0.972&
&0.996&0.964&0.979&0.995&0.983&0.983
\\\midrule
EWC-On
&0.971&0.994&0.934&0.982&0.932&0.963&
&0.906&0.966&0.967&0.965&0.889&0.938&
&0.758&0.804&0.803&0.952&0.960&0.855
\\
LwF
&0.998&0.979&0.997&\textbf{0.999}&0.985&0.992&
&0.974&0.928&0.863&0.832&0.513&0.822&
&0.948&0.873&0.671&0.505&0.514&0.702
\\
HAT
&0.999&\textbf{0.996}&0.999&0.998&0.990&\textbf{0.997}&
&0.971&0.967&0.970&0.976&0.924&0.962&
&0.988&0.911&\textbf{0.953}&\textbf{0.985}&0.977&0.963
\\\midrule
\textbf{ours}
&\textbf{1.00}&0.994&\textbf{1.00}&\textbf{0.999}&\textbf{0.993}&\textbf{0.997}&
&\textbf{0.978}&\textbf{0.972}&\textbf{0.983}&\textbf{0.988}&\textbf{0.946}&\textbf{0.974}&
&\textbf{0.994}&\textbf{0.917}&0.950&0.983&\textbf{0.978}&\textbf{0.964}
\\\bottomrule
\end{tabular}}
\caption{Task-incremental results. For each method, we report the final accuracy on all task after incremental training.}
\label{tab:task_oracle}
\end{table*}

%% file: floats/episodic_memory_plots.tex
\begin{figure}[t]
\centering
\resizebox{\columnwidth}{!}{
\begin{tabular}{cc}
\includegraphics[width=\columnwidth]{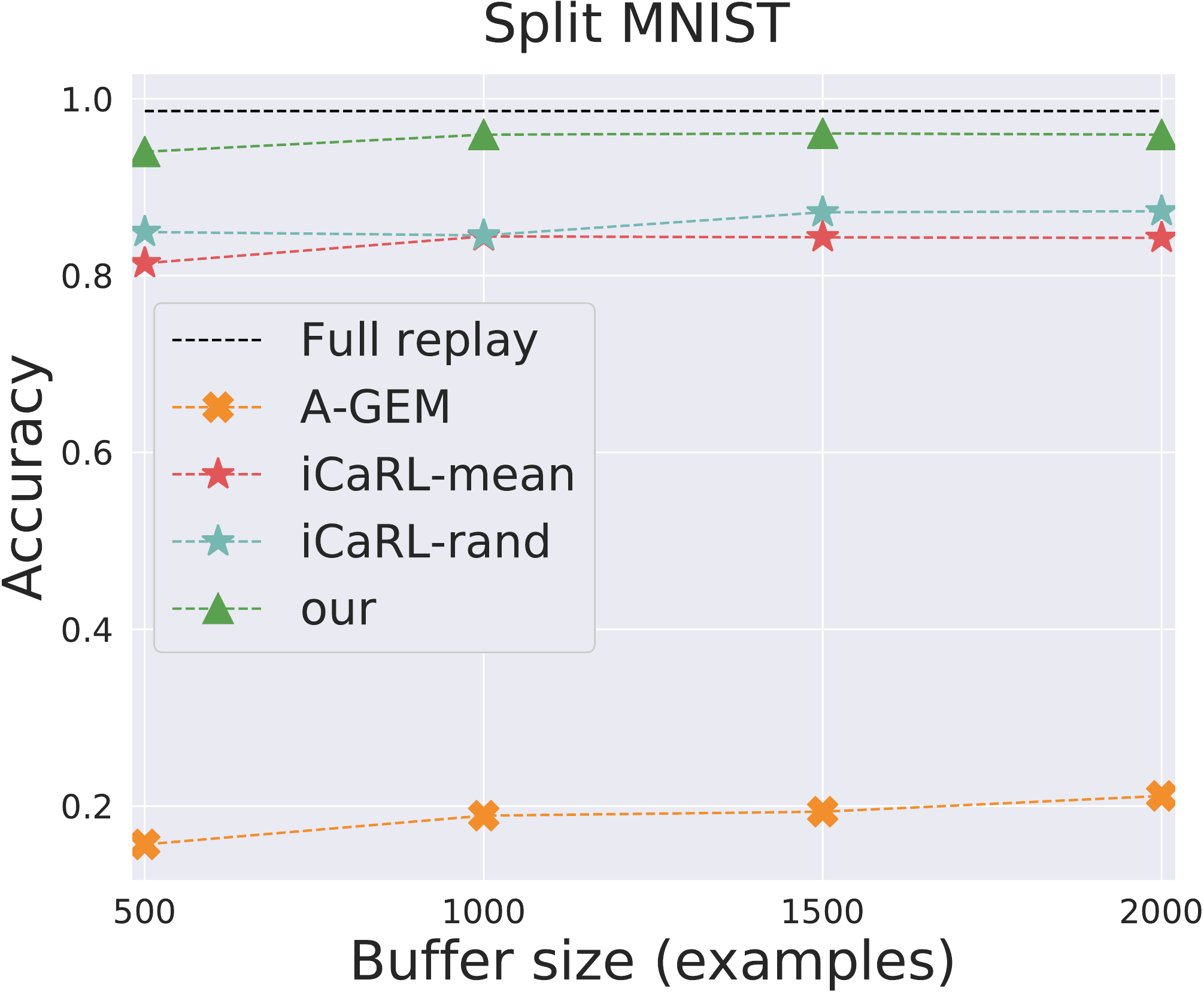}&
\includegraphics[width=0.96\columnwidth]{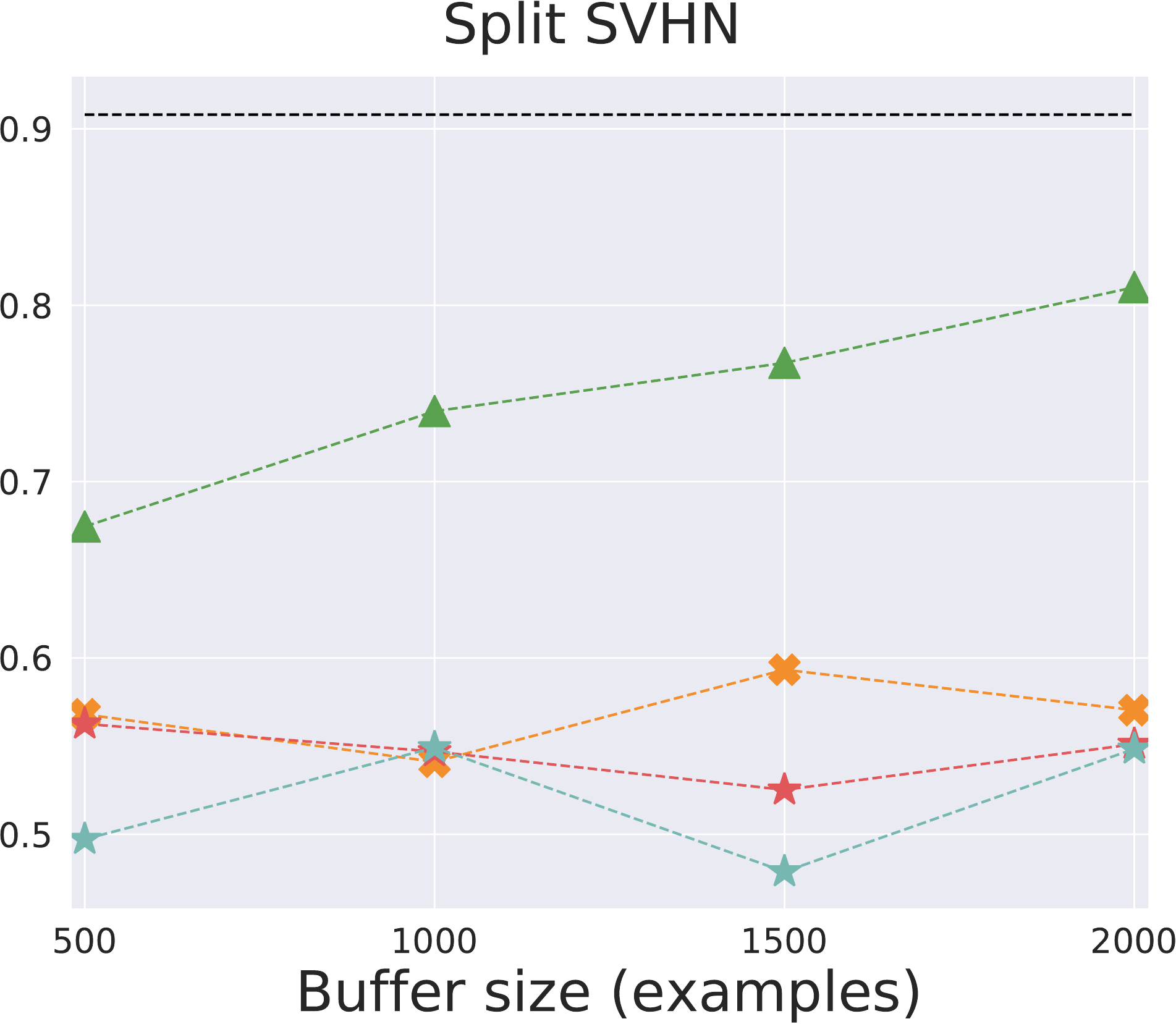}
\end{tabular}}
\caption{Final mean accuracy on all tasks when an episodic memory is employed, as a function of the buffer capacity.}
\label{fig:episodic memory_plots}
\end{figure}

%% file: floats/generative_memory_table.tex
\begin{table}[t]
\centering
\resizebox{\columnwidth}{!}{
\begin{tabular}{lcccc}
\toprule
& MNIST & SVHN & CIFAR-10 & Imagenet-50\\\midrule
DGMw~\cite{ostapenko2019learning}&0.9646&0.7438&0.5621&0.1782\\
DGMa~\cite{ostapenko2019learning}&\textbf{0.9792}&0.6689&0.5175&0.1516\\\midrule
ours &0.9727&\textbf{0.8341}&\textbf{0.7006}&\textbf{0.3524}\\
\bottomrule
\end{tabular}}
\caption{Class-incremental continual learning results, when replayed examples are provided by a generative model.}
\label{tab:generative_replay}
\end{table}

%% file: floats/memory_consumption_figure.tex
\begin{figure}[b]
    \centering
    \includegraphics[width=0.8\columnwidth]{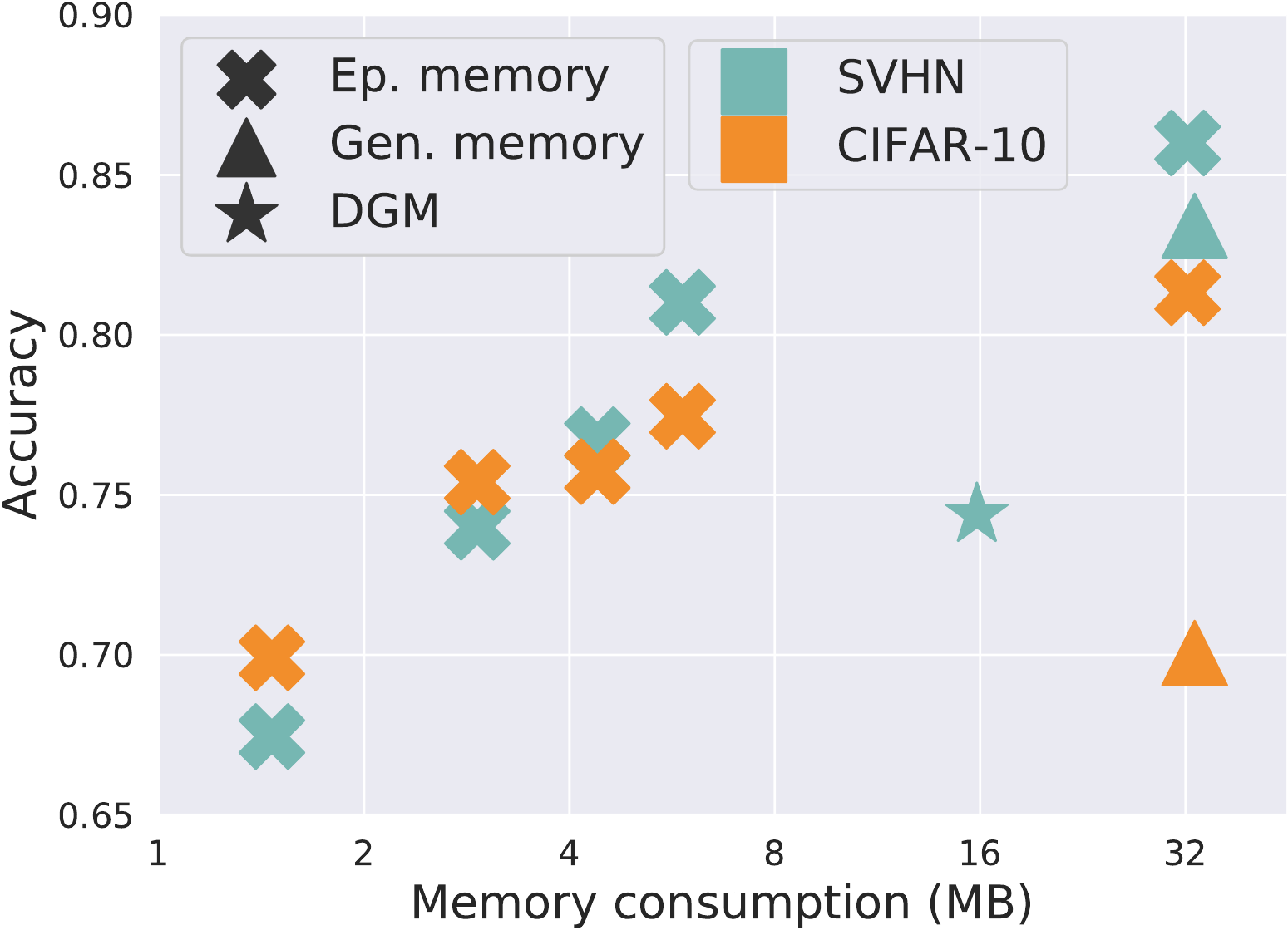}
    \caption{Accuracy as a function of replay memory budget.}
    \label{fig:memory_consumption}
\end{figure}

%% file: floats/gates_visualization_figure.tex
\begin{figure*}[bth]
\begin{tabular}{cc}
\centering
\includegraphics[width=\columnwidth]{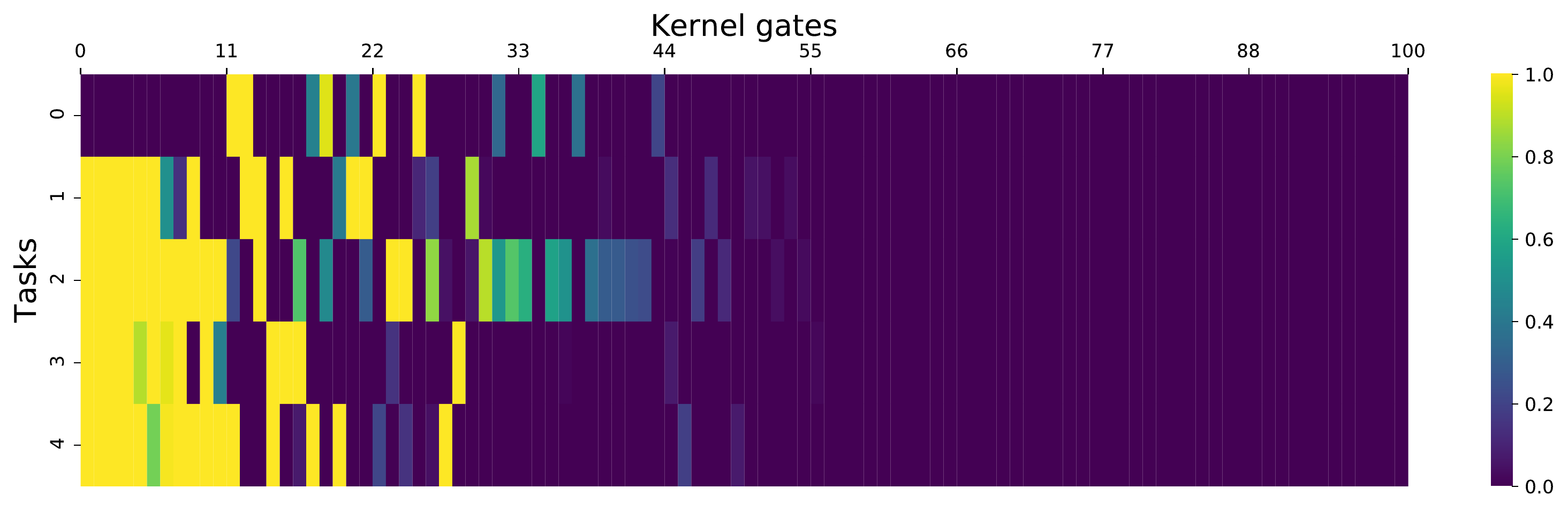}&
\includegraphics[width=\columnwidth]{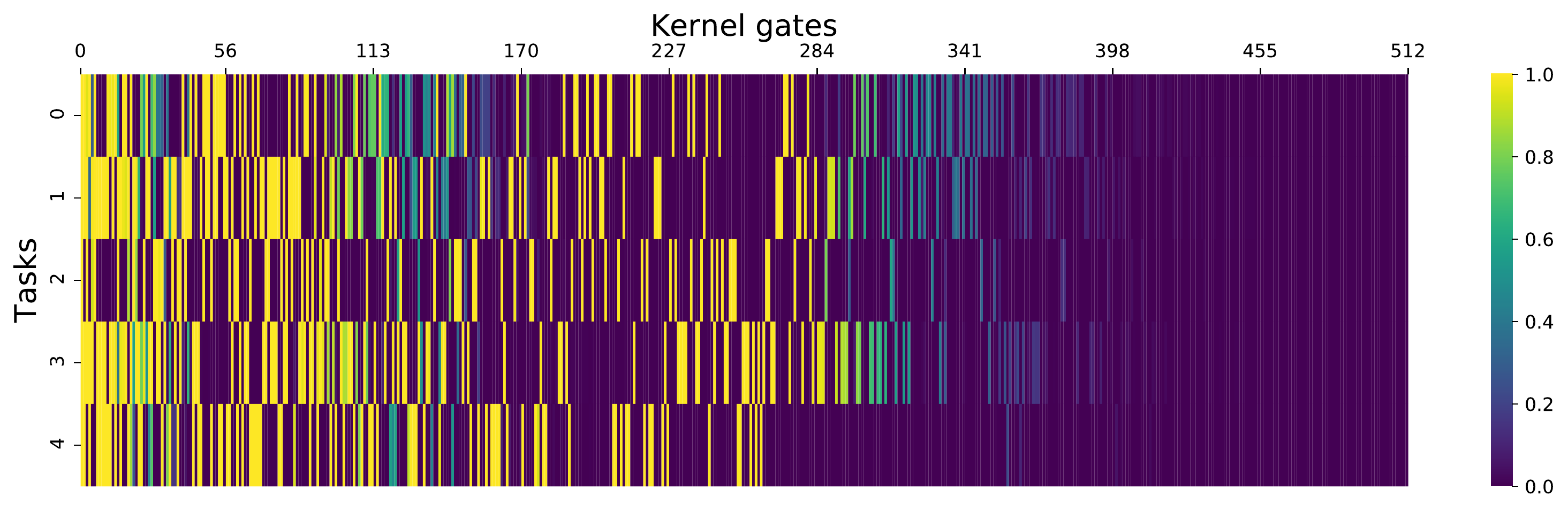}\\
\includegraphics[width=\columnwidth]{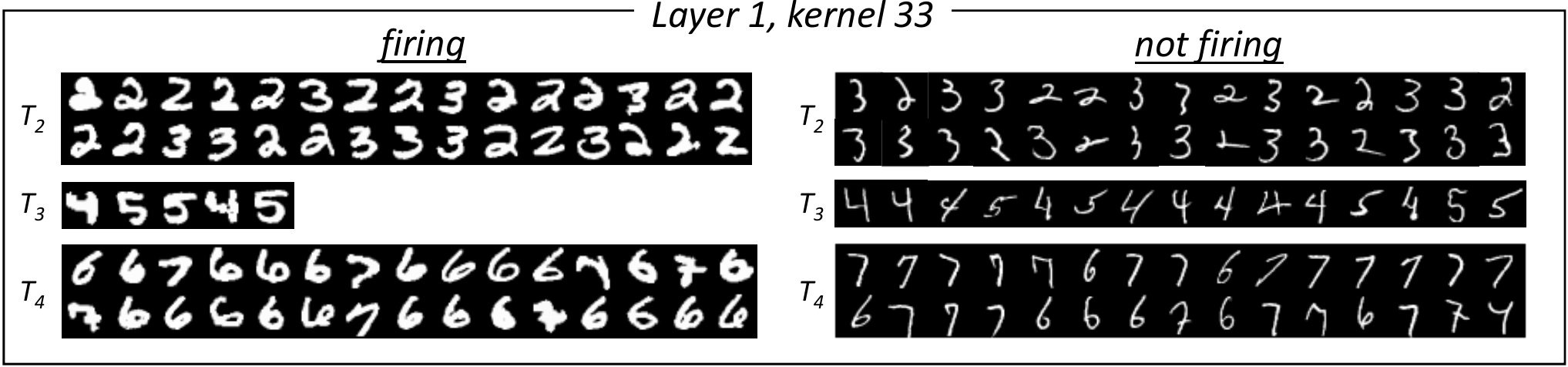}&
\includegraphics[width=\columnwidth]{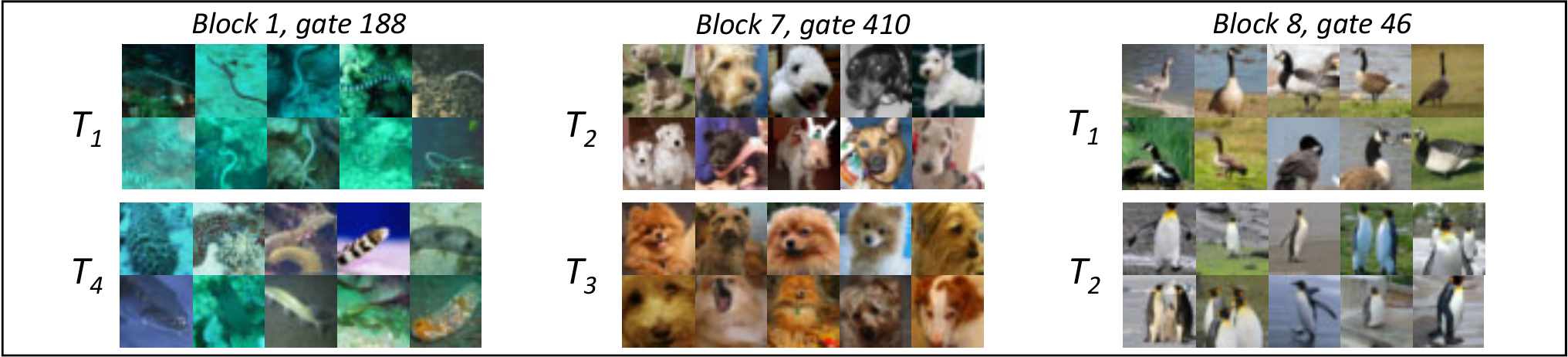}
\end{tabular}
\caption{Illustration of the gate execution patterns for continually trained models on MNIST (left) and Imagenet-50 (right) datasets. The histograms in the top left and top right show the firing probability of gates in the 1st layer and the 5th residual block respectively. For better illustration, gates are sorted by overall execution rate over all tasks. The bottom-left box shows images from different tasks either triggering or not triggering a specific gate on Split MNIST. The bottom-right box illustrates how - on Imagenet-50 - correlated classes from different tasks fire the same gates (e.g., fishes, different breeds of dogs, birds).}
\label{fig:gates_visualization}
\end{figure*}

%% file: floats/mac_count_table.tex
\begin{table}[b]
\centering
\resizebox{\columnwidth}{!}{
\begin{tabular}{lcccccc}
\toprule
 & \multicolumn{3}{c}{Split MNIST} & \multicolumn{3}{c}{Split CIFAR-10}\\
 & \multicolumn{3}{c}{\textit{\small{(Simple CNN)}}} & \multicolumn{3}{c}{\textit{\small{(ResNet-18)}}}\\\midrule
 & HAT & our & our & HAT & our & our\\
 & \small{TI} & \small{TI} & \small{CI} & \small{TI} & \small{TI} & \small{CI}\\\midrule
Up to $T_1$ & 0.151 & 0.064 & 0.064 & 31.937 & 2.650 & 2.650\\
Up to $T_2$ & 0.168 & 0.101 & 0.209 & 32.234 & 4.628 & 9.199\\
Up to $T_3$ & 0.194 & 0.137 & 0.428 & 36.328 & 5.028 & 15.024\\
Up to $T_4$ & 0.221 & 0.136 & 0.559 & 38.040 & 5.181 & 20.680\\
Up to $T_5$ & 0.240 & 0.142 & 0.725 & 39.835 & 5.005 & 24.927\\\midrule
backbone & \multicolumn{3}{c}{0.926} & \multicolumn{3}{c}{479.920}\\\bottomrule
\end{tabular}}
\caption{Average MAC counts ($\times 10^6$) of inference in Split MNIST and Split CIFAR-10. We compute MACs on the test sets, at different stages of the optimization (up to $T_t$), both in task-incremental (TI) and class-incremental (CI) setups.}
\label{tab:mac_counts}
\end{table}

%% file: sections/conclusion.tex
\section{Conclusions}
We presented a novel framework based on conditional computation to tackle catastrophic forgetting in convolutional neural networks.
Having task-specific light-weight gating modules allows us to prevent catastrophic forgetting of previously learned knowledge.
Besides learning new features for new tasks, the gates allow for dynamic usage of previously learned knowledge to improve performance.
Our method can be employed both in the presence and in the absence of task labels during test.
In the latter case, a task classifier is trained to take the place of a task oracle.
Through extensive experiments, we validated the performance of our model against existing methods both in task-incremental and class-incremental settings and demonstrated state-of-the-art results in four continual learning datasets.

%% file: supplementary/sections/implementation_details.tex
\section{Training details and hyperparameters}
\input{supplementary/floats/hyperparameters_table.tex}
In this section we report training details and hyperparameters used for the optimization of our model. As already specified in Sec.~\ref{sec:datasets} of the main paper, all models were trained with Stochastic Gradient Descent with momentum. Gradient clipping was utilized, ensuring the gradient magnitude to be lower than a predetermined threshold. Moreover, we employed a scheduler dividing the learning rate by a factor of $10$ at certain epochs. Such details can be found, for each dataset, in Tab.~\ref{tab:hyperparameters}, where we highlighted two sets of hyperparameters:
\begin{itemize}
    \item \textit{optim}: general optimization choices that were kept fixed both for our model and competing methods, in order to ensure fairness.
    \item \textit{our}: hyperparameters that only concern our model, such as the weight of the sparsity loss and the number of epochs after which sparsity was introduced (patience).
\end{itemize}

%% file: supplementary/floats/hyperparameters_table.tex
\begin{table}[b]
\centering
\begin{tabular}{c|lcc}
\toprule
\multicolumn{1}{c}{}&&Split MNIST&Split SVHN\\\midrule
\multirow{7}{*}{\rotatebox[origin=c]{90}{\textit{optim}}}
&batch size&$256$&$256$\\
&learning rate&$0.01$&$0.01$\\
&momentum&$0.9$&$0.9$\\
&lr decay&-&$[400,600]$\\
&weight decay&$5e-4$&$5e-4$\\
&epochs per task&$400$&$800$\\
&grad. clip&$1$&$1$\\\midrule
\multirow{2}{*}{\rotatebox[origin=c]{90}{\textit{our}}}
&$\lambda_s$&$0.5$&$0.5$\\
&$\mathcal{L}_{sparse}$ patience&$20$&$20$\\
\midrule\midrule
\multicolumn{1}{c}{}&&Split CIFAR-10&Imagenet-50\\\midrule
\multirow{7}{*}{\rotatebox[origin=c]{90}{\textit{optim}}}
&batch size&$64$&$64$\\
&learning rate&$0.1$&$0.1$\\
&momentum&$0.9$&$0.9$\\
&lr decay&$[100,150]$&$[100,150]$\\
&weight decay&$5e-4$&$5e-4$\\
&epochs per task&$200$&$200$\\
&grad. clip&$1$&$1$\\\midrule
\multirow{2}{*}{\rotatebox[origin=c]{90}{\textit{our}}}
&$\lambda_s$&$1$&$1$\\
&$\mathcal{L}_{sparse}$ patience&$10$&$0$\\
\bottomrule
\end{tabular}
\caption{Hyperparameters table.}
\label{tab:hyperparameters}
\end{table}

%% file: supplementary/sections/wgan_details.tex
\section{WGAN details}
This section illustrates architectures and training details for the generative models employed in Sec.~\ref{sec:generative_memory} of the main paper. As stated in the manuscript, we rely on the framework of Wasserstein GANs with Gradient Penalty (WGAN-GP,~\cite{wgangp}). The reader can find the specification of the architecture in Tab.~\ref{tab:wgan_arch}. For every dataset, we trained the WGANs for $2\times10^5$ total iterations, each of which was composed by 5 and 1 discriminator and generator updates respectively. As for the optimization, we rely on Adam~\cite{adam} with a learning rate of $10^{-4}$, fixing $\beta_1=0.5$ and $\beta_2=0.9$. The batch size was set to 64. The weight for gradient penalty~\cite{wgangp} was set to 10. Inputs were normalized before being fed to the discriminator. Specifically, for MNIST we normalize each image into the range $[0, 1]$, whilst for other datasets we map inputs into the range $[-1, 1]$.
\subsection{On mixing real and fake images for rehearsal.}
\label{sec:task_flaw}
\input{supplementary/floats/task_classifier_flaw_figure.tex}
The common practice when adopting generative replay for continual learning is to exploit a generative model to synthesize examples for prior tasks $\{1,\dots,t-1\}$, while utilizing real examples as representative of the current task $t$. 
In early experiments we followed this exact approach, but it led to sub-optimal results.
Indeed, the task classifier consistently reached good discrimination capabilities during training, yielding very poor performances at test time.
After an in-depth analysis, we conjectured that the task classifier, while being trained on a mixture of real and fake examples, fell into the following very poor classification logic (Fig.~\ref{fig:task_classifier_flaw}).
It first discriminated between the nature of the image (real/fake), learning to map real examples to task $t$.
Only for inputs deemed as fake, a further categorization into tasks $\{1,\dots,t-1\}$ was carried out. Such a behavior, perfectly legit during training, led to terrible test performances. 
Indeed, during test only real examples are presented to the network, causing the task classifier to consistently label them as coming from task $t$.

To overcome such an issue, we remove mixing of real and fake examples during rehearsal, by presenting to the task classifier fake examples also for the task $t$. In the incremental learning paradigm, this only requires to shift the training of the WGAN generators from the end of a given task to its beginning.

%% file: supplementary/floats/task_classifier_flaw_figure.tex
\begin{figure}[b]
\centering
\includegraphics[width=\columnwidth]{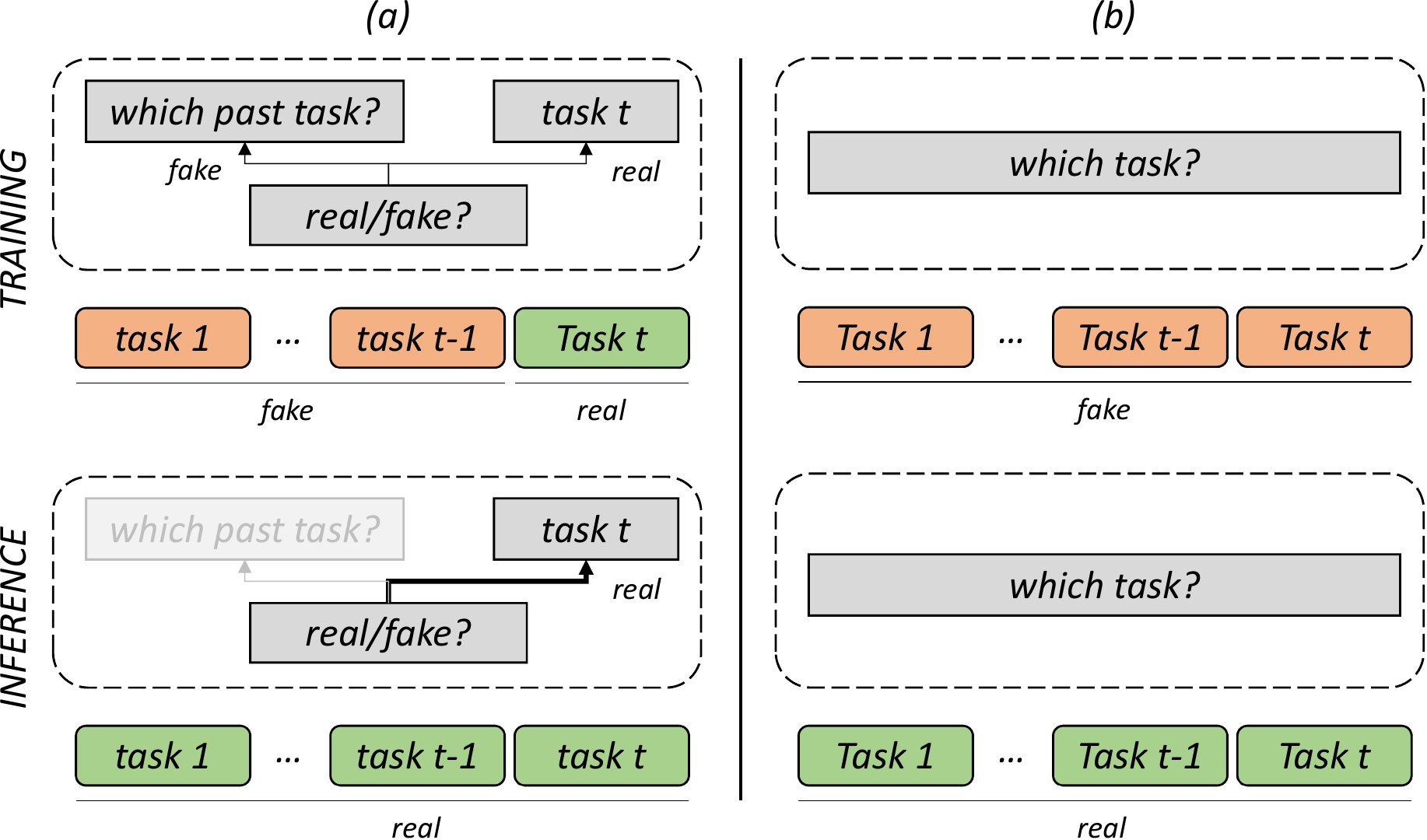}
\caption{Illustration of (a) the degenerate behavior of the task classifier when rehearsed with a mix of real and generated examples and (b) the proposed solution. See Sec~\ref{sec:task_flaw} for details.}
\label{fig:task_classifier_flaw}
\end{figure}

%% file: supplementary/sections/numbers_for_figures.tex
\input{supplementary/floats/episodic_memory_table.tex}
\input{supplementary/floats/memory_consumption_table.tex}
\section{Quantitative results for figures}
To foster future comparisons with our work, we report in this section quantitative results that are represented in Fig.~\ref{fig:episodic memory_plots} and~\ref{fig:memory_consumption} of the main paper.
Such quantities can be found in Tab.~\ref{tab:episodic memory_plots} and~\ref{tab:memory_consumption} respectively.

%% file: supplementary/floats/episodic_memory_table.tex
\begin{table}[t]
\centering
\resizebox{\columnwidth}{!}{
\begin{tabular}{c|lcccc}
\toprule
\multicolumn{1}{c}{}&&$C=500$&$C=1000$&$C=1500$&$C=2000$\\\midrule
\multirow{5}{*}{\rotatebox[origin=c]{90}{MNIST}}
&Full Replay&0.9861&0.9861&0.9861&0.9861\\
&A-GEM~\cite{agem}&0.1567&0.1892&0.1937&0.2115\\
&iCaRL-rand~\cite{icarl}&0.8493&0.8455&0.8716&0.8728\\
&iCaRL-mean~\cite{icarl}&0.8140&0.8443&0.8433&0.8426\\
\cmidrule{2-6}
&ours&\textbf{0.9401}&\textbf{0.9594}&\textbf{0.9608}&\textbf{0.9594}\\\midrule
\multirow{5}{*}{\rotatebox[origin=c]{90}{SVHN}}
&Full Replay&0.9081&0.9081&0.9081&0.9081\\
&A-GEM~\cite{agem}&0.5680&0.5411&0.5933&0.5704\\
&iCaRL-rand~\cite{icarl}&0.4972&0.5492&0.4788&0.5484\\
&iCaRL-mean~\cite{icarl}&0.5626&0.5469&0.5252&0.5511\\\cmidrule{2-6}
&ours&\textbf{0.6745}&\textbf{0.7399}&\textbf{0.7673}&\textbf{0.8102}\\\bottomrule
\end{tabular}}
\caption{Numerical results for Fig.~\ref{fig:episodic memory_plots} in the main paper. Average accuracy for the episodic memory experiment, for different buffer sizes ($C$).}
\label{tab:episodic memory_plots}
\end{table}

%% file: supplementary/floats/memory_consumption_table.tex
\begin{table}[b]
\centering
\begin{tabular}{c|lcccc}
\toprule
\multicolumn{1}{c}{}&&\multicolumn{2}{c}{SVHN}&\multicolumn{2}{c}{CIFAR-10}\\
\multicolumn{2}{c}{}&\textit{Acc.}&MB&\textit{Acc.}&MB\\\midrule
\multirow{4}{*}{\rotatebox[origin=c]{90}{\textit{episodic}}}
&Em1 & 0.6745 & 1.46 & 0.6991 & 1.46\\
&Em2 & 0.7399 & 2.93 & 0.7540 & 2.93\\
&Em3 & 0.7673 & 4.39 & 0.7573 & 4.39\\
&Em4 & 0.8102 & 5.86 & 0.7746 & 5.86\\
&Em5 & 0.8600 & 32.22 & 0.8132 & 32.22\\\midrule
\multirow{2}{*}{\rotatebox[origin=c]{90}{\textit{gen.}}}
&DGM~\cite{ostapenko2019learning} & 0.7438 & 15.82 & - & -\\
&Gm1 & 0.8341 & 33.00 & 0.7006 & 33.00\\\bottomrule
\end{tabular}
\caption{Numerical values for the memory consumption experiment represented in Fig.~\ref{fig:memory_consumption} of the main paper.}
\label{tab:memory_consumption}
\end{table}

%% file: supplementary/sections/conditional_generators.tex
\section{Comparison w.r.t. conditional generators}
\input{supplementary/floats/gen_class_vs_task_table.tex}
To validate the beneficial effect of the employment of generated examples for the rehearsal of task prediction only, we compare our model based on generative memory (Sec.~\ref{sec:generative_memory} of the main paper) against a further baseline. To this end, we still train a WGAN-GP for each task, but instead of training \textit{unconditional} models we train \textit{class-conditional} ones, following the AC-GAN framework~\cite{acgan}. After training $N$ conditional generators, we train the backbone model by generating labeled examples in an i.i.d fashion. We refer to this baseline as C-Gen, and report the final results in Tab.~\ref{tab:gen_class_vs_task}. The results presented for Split SVHN and Split CIFAR-10, illustrate that generative rehearsal at a task level, instead of at a class level, is beneficial in both datasets. We believe our method behaves better for two reasons.
First, our model never updates classification heads guided by a loss function computed on generated examples (i.e., potentially poor in visual quality). Therefore, when the task label gets predicted correctly, the classification accuracy is comparable to the one achieved in a task-incremental setup. 
Moreover, given equivalent generator capacities, conditional generative modeling may be more complex than unconditional modeling, potentially resulting in higher degradation of generated examples.

%% file: supplementary/floats/gen_class_vs_task_table.tex
\begin{table}[b]
\centering
\resizebox{\columnwidth}{!}{
\begin{tabular}{lcccc}
\toprule
 & \makecell{class\\conditioning} & \makecell{rehearsal\\level} & SVHN & CIFAR-10\\\midrule
C-Gen & \cmark & class & 0.7847 & 0.6384\\\bottomrule
ours & \xmark & task & \textbf{0.8341} & \textbf{0.7006}\\
\end{tabular}}
\caption{Performance of our model based on generative memory against a baseline comprising a class-conditional generator for each task (C-Gen). }
\label{tab:gen_class_vs_task}
\end{table}

%% file: supplementary/sections/reproducibility.tex
\section{Confidence of task-incremental results}
To validate the gap between our model's performance with respect to HAT (Tab.~\ref{tab:task_oracle} in the main paper), we report the confidence of such experiment by repeating it 5 times with different random seeds.
Results in Tab.~\ref{tab:repro} show that the margin between our proposal and HAT is slight, yet consistent.
\input{supplementary/floats/reproducibility_hat}

%% file: supplementary/floats/reproducibility_hat.tex
\begin{table}[bth]
\centering
\resizebox{\columnwidth}{!}{
\begin{tabular}{cccc}
\toprule
& MNIST & SVHN & CIFAR-10\\
\midrule
HAT & \res{0.997}{4.00\mathrm{e}{-4}} & \res{0.964}{1.72\mathrm{e}{-3}} & \res{0.964}{1.20\mathrm{e}{-3}}\\
our & \res{0.998}{4.89\mathrm{e}{-4}} & \res{0.974}{4.00\mathrm{e}{-4}} & \res{0.966}{1.67\mathrm{e}{-3}}\\
\bottomrule
\end{tabular}}
\caption{Task-IL results averaged across 5 runs.}
\vspace{-0.45cm}
\label{tab:repro}
\end{table}

%% file: supplementary/floats/wgan_table.tex
\begin{table*}[t]
\centering
\begin{tabular}{lccc}
\toprule
&& Generator & Discriminator\\\midrule
\makecell{Split MNIST}&&
\makecell{% MNIST Generator
Linear(128,4096)\\
ReLU\\
Reshape(256,4,4)\\
ConvTranspose2d(256,128,ks=(5,5))\\
ReLU\\
ConvTranspose2d(128, 64, ks=(5,5))\\
ReLU\\
ConvTranspose2d(64, 1, ks=(8,8), s=(2,2))\\
Sigmoid
}
&
\makecell{% MNIST Discriminator
Conv2d(1,64,ks=(5,5),s=(2, 2))\\
ReLU\\
Conv2d(64,128,ks=(5,5),s=(2, 2))\\
ReLU\\
Conv2d(64,128,ks=(5,5),s=(2,2))\\
ReLU\\
Flatten\\
Linear(4096,1)
}
\\\midrule
\makecell{Split SVHN}&&
\makecell{% SVHN Generator
Linear(128,8192)\\
BatchNorm1d\\
ReLU\\
Reshape(512,4,4)\\
ConvTranspose2d(512,256,ks=(2,2))\\
BatchNorm2d\\
ReLU\\
ConvTranspose2d(256, 128, ks=(2,2))\\
BatchNorm2d\\
ReLU\\
ConvTranspose2d(128, 3, ks=(2,2), s=(2,2))\\
TanH
}
&
\makecell{% SVHN Discriminator
Conv2d(3,128,ks=(3,3),s=(2,2))\\
LeakyReLU(ns=0.01)\\
Conv2d(128,256,ks=(3,3),s=(2,2))\\
LeakyReLU(ns=0.01)\\
Conv2d(256,512,ks=(3,3),s=(2,2))\\
LeakyReLU(ns=0.01)\\
Flatten\\
Linear(8192,1)
}
\\\midrule
\makecell{Split CIFAR-10}&&
\makecell{% CIFAR-10 Generator
Linear(128,8192)\\
BatchNorm1d\\
ReLU\\
Reshape(512,4,4)\\
ConvTranspose2d(512,256,ks=(2,2))\\
BatchNorm2d\\
ReLU\\
ConvTranspose2d(256, 128, ks=(2,2))\\
BatchNorm2d\\
ReLU\\
ConvTranspose2d(128, 3, ks=(2,2), s=(2,2))\\
TanH
}
&
\makecell{% CIFAR-10 Discriminator
Conv2d(3,128,ks=(3,3),s=(2,2))\\
LeakyReLU(ns=0.01)\\
Conv2d(128,256,ks=(3,3),s=(2,2))\\
LeakyReLU(ns=0.01)\\
Conv2d(256,512,ks=(3,3),s=(2,2))\\
LeakyReLU(ns=0.01)\\
Flatten\\
Linear(8192,1)
}
\\\midrule
\makecell{Imagenet-50}&&
\makecell{% Imagenet-50 Generator
Linear(128,8192)\\
BatchNorm1d\\
ReLU\\
Reshape(512,4,4)\\
ConvTranspose2d(512,256,ks=(2,2))\\
BatchNorm2d\\
ReLU\\
ConvTranspose2d(256, 128, ks=(2,2))\\
BatchNorm2d\\
ReLU\\
ConvTranspose2d(128, 3, ks=(2,2), s=(2,2))\\
TanH
}
&
\makecell{% Imagenet-50 Discriminator
Conv2d(3,128,ks=(3,3),s=(2,2))\\
LeakyReLU(ns=0.01)\\
Conv2d(128,256,ks=(3,3),s=(2,2))\\
LeakyReLU(ns=0.01)\\
Conv2d(256,512,ks=(3,3),s=(2,2))\\
LeakyReLU(ns=0.01)\\
Flatten\\
Linear(8192,1)
}
\\\bottomrule
\end{tabular}
\caption{Architecture of the WGAN employed for the generative experiment. In the table, \textit{ks} indicates kernel sizes, \textit{s} identifies strides, and $ns$ refers to the negative slope of Leaky ReLU activations.}
\label{tab:wgan_arch}
\end{table*}